\documentclass[10pt,twocolumn,letterpaper]{article}

\usepackage{cvpr}              % To produce the CAMERA-READY version
\usepackage{graphicx}
\usepackage{amsmath}
\usepackage{amssymb}
\usepackage{booktabs}
\usepackage{float}
\usepackage{acronym}
\usepackage{xcolor}

\acrodef{SOD}[SOD]{\emph{Salient object detection}}
\acrodef{ROI}[ROI]{Region of Interest}
\acrodef{CNN}[CNN]{Convolutional Neural Network}
\acrodef{mAP}[mAP]{mean Average Precision}
\acrodef{NFO}[NFO]{Non-subject Foreground Object}
\acrodef{DETR}[DETR]{DEtection TRansformer}
\usepackage[pagebackref,breaklinks,colorlinks]{hyperref}

\usepackage[capitalize]{cleveref}
\crefname{section}{Sec.}{Secs.}
\Crefname{section}{Section}{Sections}
\Crefname{table}{Table}{Tables}
\crefname{table}{Tab.}{Tabs.}

\def\cvprPaperID{****} % *** Enter the CVPR Paper ID here
\def\confName{CVPR}
\def\confYear{2022}

\begin{document}

\title{ ImageSubject: A Large-scale Dataset for Subject Detection}

% \author{Xin Miao, Huayan Wang, Jun Fu, Jiayi Liu, Shen Wang, Zhenyu Liao}
% \affiliations{Kuaishou Technology\\ xinmiao@kuaishou.com, wanghuayan@kuaishou.com, fujun@kuaishou.com, 
%              jiayiliu@kuaishou.com,  wangshen@kuaishou.com, liaozhenyu2004@gmail.com}

\author{Xin Miao\\
Kuaishou Technology\\
{\tt\small xin.miao@mavs.uta.edu}
\and
Jiayi Liu\\
Kuaishou Technology\\
{\tt\small jiayi.uiuc@gmail.com}
\and
Jun Fu\\
Kuaishou Technology\\
{\tt\small fujun@kuaishou.com}
\and
Huayan Wang\\
Kuaishou Technology\\
{\tt\small uayanw@cs.stanford.edu}
}
\maketitle

%%%%%%%%% ABSTRACT
\begin{abstract}
Main subjects usually exist in the images or videos, as they are the objects that the photographer wants to highlight. Human viewers can easily identify them but algorithms often confuse them with other objects. Detecting the main subjects is an important technique to help machines understand the content of images and videos. We present a new dataset with the goal of training models to understand the layout of the objects and the context of the image then to find the main subjects among them. This is achieved in three aspects.  By gathering images from movie shots created by directors with professional shooting skills, we collect the dataset with strong diversity, specifically, it contains 107\,700 images from 21\,540 movie shots.  We labeled them with the bounding box labels for two classes:  subject and non-subject foreground object. We present a detailed analysis of the dataset and compare the task with saliency detection and object detection. ImageSubject is the first dataset that tries to localize the subject in an image that the photographer wants to highlight. Moreover, we find the transformer-based detection model offers the best result among other popular model architectures. Finally, we discuss the potential applications and conclude with the importance of the dataset.

\end{abstract}

% TODO other exploration direction - subtitle vs image interaction

\section{Introduction}
\label{sec:intro}

\begin{figure}[t]
\begin{center}
\includegraphics[width=1\linewidth]{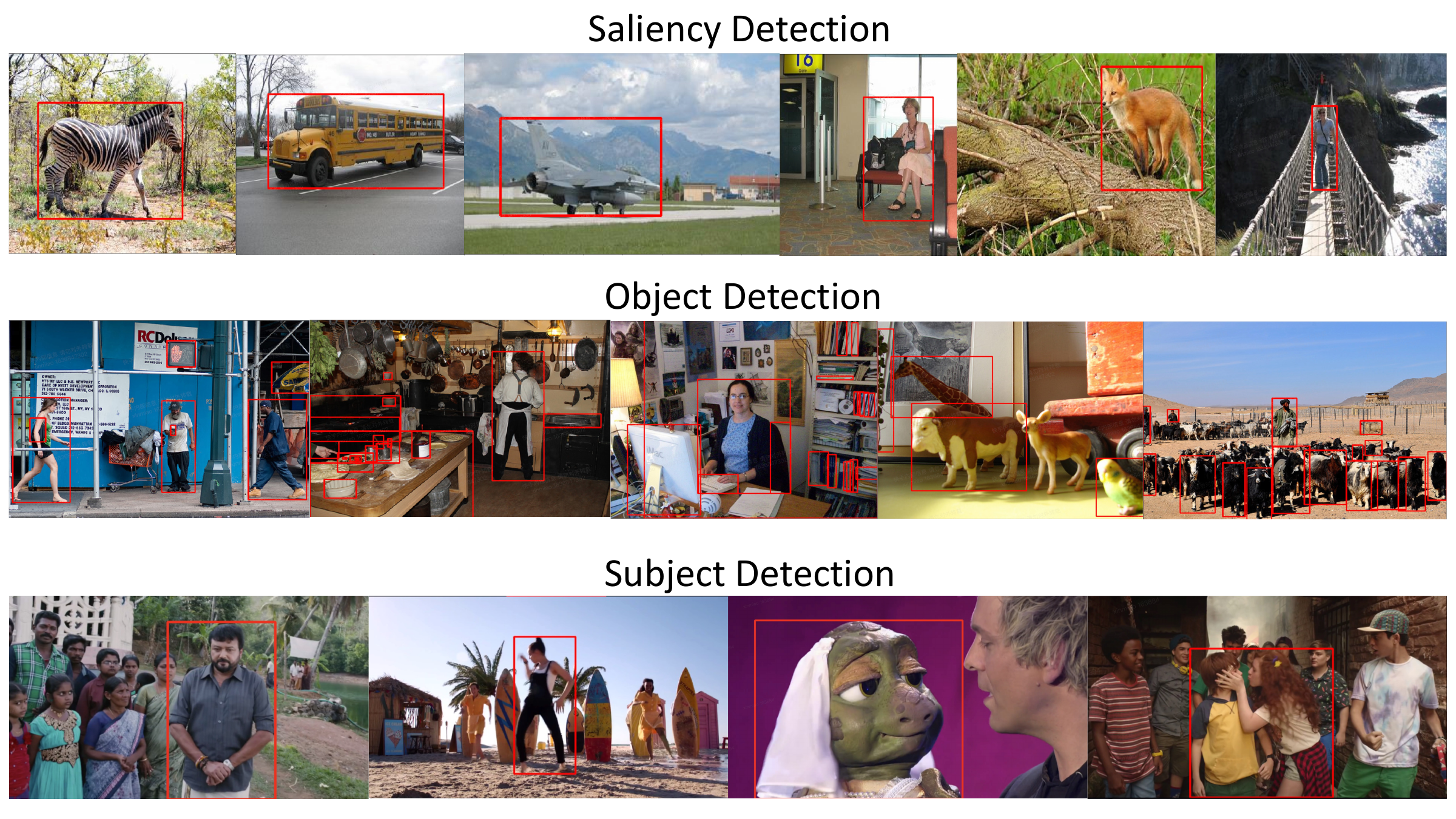}
\end{center}
\vspace{-0.3cm}
\caption{Compare our proposed subject detection dataset to the most relevant datasets for object detection and saliency detection. The saliency detection dataset (first row) usually only contain a few objects for each image, the model learns to segment the object from background instead of understanding the complex context. The object detection dataset (second row) label all the defined objects in an image, without differentiating the importance between them.  Our subject detection (third row)  label the subject that the photographer wants the highlight when the image was taken.}
\label{fig:figure1}
\end{figure}

Detecting the most important regions or object of an image has recently drawn significant attention in computer vision \cite{miao2018direct, miao2020efficient, miao2019deep, zheng2018fast, gu2018asynchronous, yue2018attentional, miao2019net} due to its fundamental role in various applications, including image editing \cite{rao2020unified},  video compression \cite{le1991mpeg}, image quality assessment \cite{wang2006modern}. \ac{SOD} defines the salient objects that attract human attention. The datasets 
\cite{wang2017learning, li2014secrets, cheng2014salientshape} collected for salient object detection usually contain a few objects (mostly one) for each image. \ac{SOD} aims at segmenting the objects from the background. There is usually no other objects besides the salient object (\cref{fig:figure1}, first row). However, in real world cases we need to distinguish the main objects based on the understanding of the scene and relations between several objects in the image or even use the temporal information from the video. So the salient objects detection model is easy to fail in these cases (see experiment part).

Web users generated billions of images or videos on social media in this web 2.0 era. Comprehending them by machines is an important topic and leads to many applications. Object detection methods \cite{redmon2016you, ren2015faster, liu2016ssd} try to localize every (known) object in the image (\cref{fig:figure1}, second row) and they detect each object independently. But most of the detected objects are not the main subjects that the user wants to emphasize. Therefore, it is a fundamental step in the image or video understanding pipeline to localize  which object is the main subject in the image.  We may design  post-processing for the object detection results to find the main subject, for example the main subject usually has the bigger scale and better clarity than environment objects. However, it is still hard to find a good rule to cover the complex situation in the subject detection task. 

Motivated by the above insights, we construct a subject detection dataset called ImageSubject in this paper (\cref{fig:figure1}, third row). We collected five key-frame images for each movie shot from the public MovieNet dataset\cite{rao2020unified}, with annotations for the bounding box for the subject in the movie shots. We define the subject as the object that has been purposely highlighted when people created the images or videos. The reason that we use movie shot as our data source is there are usually subjects that the directors want to highlight when they make a movie shot.  It is more consistent to label the subjects afterward using general viewers to annotate. Since five key frames are extracted from each shot and then labeled, so not only the spatial information but also the temporal information are included in the dataset.

We further conduct detailed analysis on ImageSubject  and compared it with other saliency detection~\cite{wang2017learning} and object detection \cite{lin2014microsoft} datasets. We did experiments for different object detection methods on our proposed ImageSubject dataset. The performance of those methods are different from their performance for object detection. Sparse-RCNN~\cite{sun2021sparse} gives the best result on COCO object detection dataset~\cite{lin2014microsoft} among these methods but \ac{DETR}~\cite{carion2020end} gives the best result on our ImageSubject. Base on the experimental results, we conclude that the transformer-based model like \ac{DETR} can exploit the relations between the objects in the image and understand the context of the whole image before proposing the final subjects bounding boxes. The state-of-the-art video object detection methods are also conducted for our dataset. We give the baseline for subject detection in this paper, and facilitate researches along this direction.

In this work, our major contributions can be summarized
in the following three aspects.
\begin{itemize}
\item[1)] To the best of our knowledge, we propose the first large-scale subject detection dataset with the professionally annotated main subject bounding box. The data is collected from movie shots created by professional directors. As the demand for understanding images and videos increases in this era, we believe our proposed dataset can help researchers develop better techniques in this direction.

\item[2)] We conduct detailed analysis for the proposed dataset and also implement experiments to show the differences between our subject detection and other tasks like saliency detection. We demonstrate that the existing datasets can not solve the proposed  subject detection task well. It is crucial to propose the subject detection dataset.

\item[3)] We try different  detection models on our new dataset include the single frame  models and temporal models for video data. We conclude that the transformer-based model, such as \ac{DETR} \cite{carion2020end},  provides better understanding of relations between objects and find the subject among them.  We also provide multiple baseline performances for subject detection.

\end{itemize}

\section{Related Work}
\label{sec:relate}

Datasets have played a critical role in the history of computer vision research. Nowadays, deep learning models usually need large-scale datasets to get good performance. More importantly, the datasets are the foundations for new research directions. The research directions related to our subject detection can be roughly split into two groups: saliency detection and object detection.

{\bf Saliency Detection} \ac{SOD} defines the salient object that attract human attention. It is an important research topic in computer vision because it is a fundamental role in various applications, e.g. image editing \cite{rao2020unified},  video compression \cite{le1991mpeg}. There are several datasets \cite{li2015visual, yan2013hierarchical, yang2013saliency, cheng2014salientshape, fan2018salient} were proposed  in the early stage, however due to the dense segmentation label is required for salient object detection, those datasets only contain limited training data. So they are unsuitable for training complex deep networks. Motivated by this observation, \cite{wang2017learning} proposed a large-scale dataset  DUTS \cite{wang2017learning} that has 10\,553 training images. MSRA10K \cite{cheng2014global} is another large-scale  \ac{SOD} dataset, it contain 10000 images with pixel accurate salient object labeling. Salient object detection aims to predict the segmentation map for the salient objects in an image. Thus the common model design for \ac{SOD} is an U-net shape network that extracts dense feature from the image. The data collected for it usually contain few objects (mostly one) for each image. It does not count the relation between different objects. Our subject detection task tries to find the main subject in the image instead of finding each object individually and predicting whether they are salient. In the experiment part, we show the \ac{SOD} model cannot detect the subject region well. Specially, the detected salient region may not be the subject that wants to be highlighted.

Another related direction is human gaze fixations prediction~\cite{wang2018salient, cornia2018predicting}.  However, limited by the data collection equipment, they usually have inadequate training samples; the gaze fixations labels are noisy; and the fixation regions are usually a small portion of objects.  Thus the datasets of gaze fixations are not suitable for the subject detection. We want to find the subjects that are purposely highlighted and convey the contextual meaning of images and videos.

{\bf Object Detection} Object detection is one of the most important research areas in computer vision. Thanks to the proposed datasets \cite{everingham2015pascal, deng2009imagenet}, especially the COCO dataset \cite{lin2014microsoft}, the detection techniques have been developed rapidly. COCO contains 80 object types and 2.5 million labeled instances in 328k images. Objects are labeled using per-instance bounding box, class and segmentation. Each image usually contains many objects and each object is considered as individual object instance. Object detection task does not consider the layout for different objects in the same image. The context understanding is not included. Object detection datasets label all the defined objects in an image, without differentiating the importance between them. Compared with it, our subject detection only focuses on the main subjects that should be highlighted in the image. We do not want to label objects as many as possible. Our dataset only contains 2 classes, the subject and the non-subject foreground object that help the model to better understand the layout and scene. And we only label the bounding boxes for them. Some subjects labeled in our ImageSubject do not belong to the defined classes in COCO dataset \cite{lin2014microsoft}.

% @jiayiliu -  展示一下我们主体识别能否识别没有见过的主题 （在物体主体里面找一找，最好不是 coco 的 label ）

Object detection models can be split into two-stages detectors~\cite{ren2015faster} and single-stage detectors ~\cite{liu2016ssd, redmon2016you}. Most of those methods predict the final bounding box depending on the initial guesses like the anchors or a grid of possible object centers. Two stage models were proposed to predict every guess region whether it has an object individually. For example, the Faster R-CNN model~\cite{ren2015faster}  predict each \ac{ROI} contains an object or not by shared parameter operations. \cite{deng2019relation} proposed to use all of the \ac{ROI} features from several frames to  augment object features of the current frame for the video object detection. One stage models usually predict the labels on the dense feature maps.

\cite{carion2020end} proposed the first transformer-based object detection framework named \ac{DETR}. In the decoder, the  object queries do the cross-attention with the dense map and the self-attention between object queries. Subject detection is a detection task too, we can take advantage of object detection models. Since we only want to predict the main subject in the image instead of all the objects. It is reasonable to use the transformer model to mining the relations between different objects  instead of predicting them independently. And the cross attention between object queries and the feature map also helps the model  understand the context of the whole image.

\begin{table}[t]
\begin{center}
\begin{tabular}{c|c|c}
\hline
Dataset & \# images   & Type \\
\hline
ImageSubject & 107\,700 & Subject detection\\
\hline
COCO \cite{lin2014microsoft}& 123\,287 &  Object detection  \\
\hline
DUTS \cite{wang2017learning} & 15\,572& Saliency detection \\
\hline
MSRA10K \cite{cheng2014global} & 10\,000& Saliency detection \\
\hline
\end{tabular}
\end{center}
\caption{Comparisons with  datasets of other tasks.}
\label{tab:table1}
\end{table}

% \begin{table*}[t]
%\begin{center}
%\begin{tabular}{c|c|c|c|c}
%\hline
%Dataset & \# images   & Subject detection& Object detection&Saliency detection\\
%\hline
%ImageSubject & 107\,700&\checkmark&&\\
%\hline
%COCO \cite{lin2014microsoft}& 123\,287 &  &\checkmark&  \\
%\hline
%DUTS \cite{wang2017learning} & 15\,572&  & &\checkmark \\
%\hline
%MSRA10K \cite{cheng2014global} & 10\,000&  & &\checkmark \\
%\hline
%\end{tabular}
%\end{center}
%\caption{Comparisons with  datasets of other tasks.}
%\label{tab:table1}
%\end{table*}

\begin{figure*}[t]
\begin{center}
\includegraphics[width=1\linewidth]{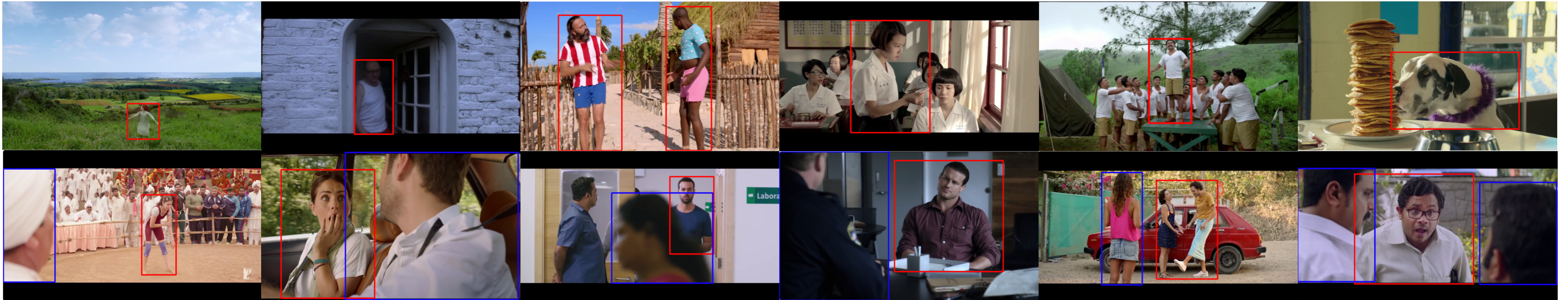}
\end{center}
\vspace{-0.3cm}
\caption{Visualization examples of our proposed ImageSubject dataset. We label the subjects in red bounding boxes (first and second rows) and label the non-subject foreground objects in blue bounding boxes (second row).}
\label{fig:figure2}
\end{figure*}

\section{Subject Detection Dataset}
\label{sec:dataset}

To facilitate the subject detection research, we collect ImageSubject containing 107\,700 images belonging to 21\,540 shots.  The details of the proposed dataset are specified as follows.

\subsection{Data Source and Categories}
We define a subject as the object in an image or video that the photographer wants to focus on and highlight, which should attract the most attention from the viewer. Especially for movies, when directors make a video shot, there is usually a main subject that they want to highlight and let the audience focus (bounding boxes in red in \cref{fig:figure2}). Thanks to the public shot type classification dataset MovieShots \cite{rao2020unified}, we use the released 33\,653 shots as our data source. We extract five key frames from each shot to be labeled. But sometimes the subjects are not foregrounded in the image, it is hard to be identified by the model, one example is the over-the-shoulder shot in movie shooting. In order to help the model better understand the scene and predict the subject precisely, we also label the non-subject foreground object (bounding boxes in blue in \cref{fig:figure2}). We define a non-subject foreground object as the object in front of the subject and it is not a subject. The non-subject foreground object is usually between the camera and the subject when the frame is made.  So there are two categories in our dataset: (1) subject (2) non-subject foreground object. ImageSubject contains layout information in a single frame and also the temporal information of a movie shot.

\subsection{Data Statistics}
The released MovieShots consists of 33\,653 shots from 7848 movies, covering a wide of movie genres. After the annotation consistency filtering, our ImageSubject dataset consists of 107\,700 images belonging to 21\,540 shots.  Most frames contain only one subject, it is consisted with our common sense that there is only one object that the director wants to focus on when make movie shots. 
Among 33\,653 shots, 6811 shots do not contain the subject, such as a picture taken for the full view of the city, we do not include these images in our dataset but will provide them for further use.  
Some frames have more than one subject are kept in the dataset. In \cref{tab:table1}, we compare ImageSubject with existing related datasets. Then the number of images in  ImageSubject is much larger than the state-of-the-art \ac{SOD} datasets DUTS and MSRA10K. Our ImageSubject dataset is the only one that can do subject detection and it is large enough that the deep network models can be trained on our data. 

\subsection{Annotation Procedure}
Thanks to the MovieShots dataset~\cite{rao2020unified}, the shots are already cleaned by removing the advertisement and text in frames. In most cases, people can identify the subject for a single image, however, in very few cases the subject can be identified based on the temporal information. So we extract five key frames from each shot and then show them to the annotators. Each annotator will see those five key frames at one time, and label the subject for each frame. At first, all the annotators are trained to label sampled examples by cinematic professionals experts, so all annotators can provide high quality labels. To ensure the high consistency, there are three rounds of annotation procedures. We remove inconsistent data by requiring at least two annotations on the same image share more than 85\% overlapping region for subjects and non-subject foreground objects.  The final bounding boxes are the average of the consistent annotations.  
We finally achieve very high consistency by removing 5302 inconsistent shots labels.

\section{Subject Detection Models}
The target of subject detection is to detect the most important object that wants to be highlighted. If there is only one object in the image, it is easy to be classified as the subject. But sometimes there are many objects in the image, the subject can be identified by its scale, position, clearness, relation with other objects and even the scene information. The \ac{SOD} models are trained to predict the segmentation map for the salient objects which is different from our task that bounding boxes need to be predicted. So we take advantage of the state-of-the-art object detection models. In \cref{sec:initial_guess_based_model} and \cref{sec:transformer_based_model}, we review the traditional initial guess based objects detection methods include one-stage and two-stage models and the recent proposed transformer-based model. Then we give a detailed analysis of why the transformer based architecture can predict the subject well.

\subsection{Initial Guess Based Model}\label{sec:initial_guess_based_model}
Generally speaking, most object detection models can be split into two categories: two-stage detectors and single-stage detectors. One of the representative two-stage works is Faster-RCNN \cite{ren2015faster}. The proposal stage pre-defines the anchors of each position on the feature and predicts its binary label (object or background). Then the \ac{ROI} module is used to extract local features and the final label is predicted by shared parameters operations. In this way, each object is predicted individually. But in subject detection, we want to contrast the scale, position, clearness, etc. information and the relations between different objects. So it is not reasonable to use this kind of method. Even there are some works \cite{hu2018relation} try to model the relations between different predictions of the proposed interest region, the local receptive field of convolutional operations still limits their performance for subject detection. Please check the experimental part for more detail. Similar to the two-stage method, most single-stage models make predictions based on the dense initial guess like anchors or possible object centers \cite{liu2016ssd, redmon2016you}. Due to the local receptive filed of \ac{CNN}, they still do not solve our task well.

\subsection{Transformer Based Model}\label{sec:transformer_based_model}
The transformer model becomes more and more popular since it was proposed \cite{vaswani2017attention}. It makes great achievements in computer vision \cite{devlin2018bert,zhang2019self} because it is non-local operation. We review the architecture of the recently proposed \ac{DETR} \cite{carion2020end} model and give the reason that it fits our task. 

Given the input image with three color channels $x \in \mathbb{R}^{3\times H_{0} \times W_{0}}$, a \ac{CNN} backbone, such as ResNet50~\cite{he2016deep}, is used first to extract a lower-resolution feature map $f \in \mathbb{R}^{C\times H \times W}$, then a $1\times1$ convolution reduces the channel dimension from $C$ to $d$, so we have the feature map $z \in \mathbb{R}^{d\times H \times W} $. There is one transformer encoder and decoder in the \ac{DETR}. Each vector at a specific position through the channel is considered as a point in the input sequence, which means that  we reshape $z$ to a feature map with the shape $d\times H W$.  Multi-head self-attention is implemented in each encoder layer. With this non-local operation, there is enough mining for the relations between all positions in the feature map. It could help the subject detection. On the other hand, due to the permutation-invariant property of transformer architecture, positional embeddings are added to each input of each attention layer. We believe the positional information is very important for subject detection because the subject is more likely to be located in the center areas. So the positional embeddings addition naturally improves the performance of subject detection.

\begin{figure}[t]
\begin{center}
\includegraphics[width=1\linewidth]{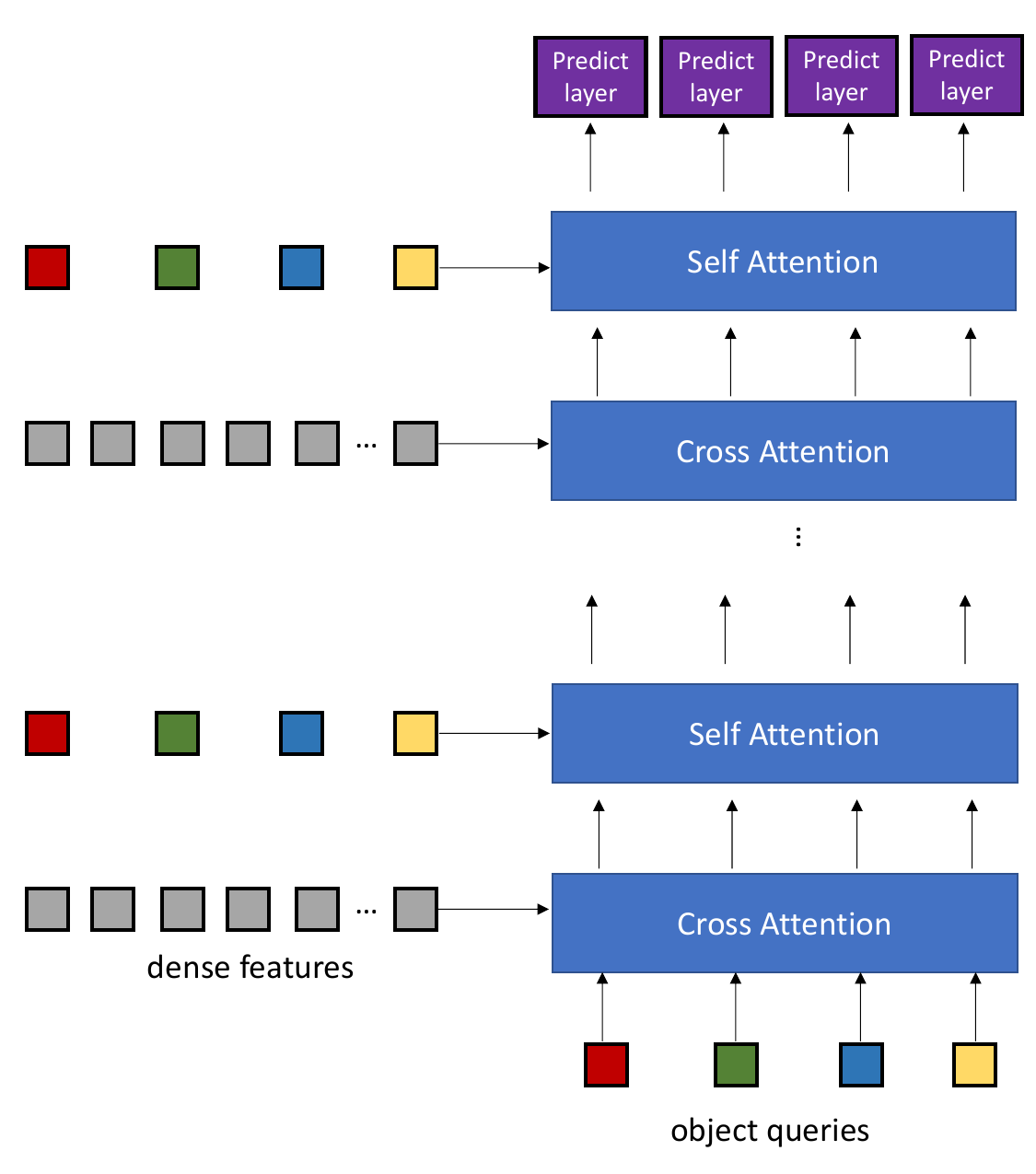}
\end{center}
\vspace{-0.3cm}
\caption{Decoder architecture of \ac{DETR}. }
\label{fig:figure3}
\end{figure}

\begin{table*}[t]
\begin{center}
\begin{tabular}{c|c|c|c|c}
\hline
& Train & Val &  Test & Total \\
\hline
\# Shots & 15\,078 & 2154 & 4308 &21\,540\\

\hline
\# Images &75\,390 &10\,770  &21\,540 & 107\,700\\
\hline
\# Subjects & 8780 & 1196 & 2575 & 12\,551\\
\hline

\# Non-subject foreground objects &76\,833 &11\,019  &21\,908& 109\,760\\
\hline
\end{tabular}
\end{center}
\caption{Statistics of our proposed ImageSubject dataset.}
\label{tab:table2}
\end{table*}

The most important module for subject detection in the \ac{DETR} architecture is the decoder. The decoder transforms $N$  embeddings of size $d$ by multi-headed attention mechanisms. The $N$ input embeddings are called object queries, they are learned positional encodings. Since the permutation-invariant property of the decoder, they are added to the input of each attention layer. As shown in \cref{fig:figure3}, on the one hand, the $N$ object queries do the cross-attention with the dense encoded feature maps from the \ac{DETR} encoder. So they can better understand the context from the whole image. As mentioned before, it is required by the subject detection task. On the other hand, the $N$ object queries also do the self-attention between themselves, so in this way the model reasons all objects using pair-wise relations between them globally. More specifically,  we can contrast the scale, position, clearness, and other information and the relations between different objects. In the experiment part, we will prove the cross-attention and the self-attention modules that no one of them can be dispensed with. The architecture of the transformer-based model \ac{DETR} naturally fits our proposed subject detection task.

After enough attention operations, feed-forward networks are used to predict the normalized center coordinates, the height and the width of the box for the $N$ objects individually. Because the average number of subjects in one image from our dataset is much less than the average number of objects in one image from the objection detection dataset, $N=100$ is enough for our task. We use the set prediction loss to train the \ac{DETR} same as its original setting.

\begin{table*}[t]
\begin{center}
\begin{tabular}{c|c|c|c|c|c|c|c|c|c}
\hline
Model& AP & AP50 &  AP75 & Subject &Non-subject  &FPS&Model type&AP on &AP on \\

&  &  &   &  &foreground && &COCO&ImageNet VID\\
\hline

RetinaNet &67.3  & 82.7 &72.2   &75.7  &58.9 &18 &Single frame&38.7&- \\
\hline
Faster-RCNN &65.8  &81.6 &72.1 &72.7&58.9 &19&Single frame&40.2&71.8\\
\hline
Sparse-RCNN (CVPR 21) & 67.7&80.8  &72.6&75.9&59.5 &17&Single frame& {\bf 45.0}&-\\
\hline
\ac{DETR}(ECCV 20) & {\bf71.4} & {\bf84.2}  & {\bf76.7} & {\bf78.1} & {\bf64.8} &22&Single frame&42.0&-\\
\hline
RDN & 63.8& - &-&69.2&58.5 &-&Temporal&-&76.7\\
\hline

MEGA &64.5 & -&-& 70.4&58.7 & -&Temporal&-&77.3\\
\hline

\end{tabular}
\end{center}
\caption{We show the prediction results by different types (single frame and temporal) detection methods on our proposed ImageSubject dataset. We use Detectron2 \cite{wu2019detectron2} to measure frame-per-second (FPS) for models on the same GPU. We also show their results on object detection datasets COCO \cite{lin2014microsoft} and ImageNet VID \cite{russakovsky2015imagenet}.}
\label{tab:tabel3}
\end{table*}

\section{Experiments}

\subsection{Data}
We conduct all the experiments on our proposed dataset ImageSubject. We split the dataset into Train, Validation (Val) and Test sets with a ratio of 7:1:2, the detail can be found in \cref{tab:table2}.

\subsection{Evaluation metrics}
Since the subject detection is a bounding box detection task. We take the commonly used \ac{mAP} as the evaluation metric. Since we only have two categories: subject and non-subject foreground object, we also show the mAP for each class. We omit \% symbol for AP in our paper. For the inference speed, we test all the methods on the same NVIDIA 2080 GPU by using the same testing code from Detectron2~\cite{wu2019detectron2}.

\subsection{Implementation Details}
We use the official implementations for \ac{DETR} \cite{carion2020end} and recent;u proposed Sparse-RCNN \cite{sun2021sparse}.
For RetinaNet \cite{lin2017focal} and Faster-RCNN \cite{ren2015faster}, we used the Detectron2 \cite{wu2019detectron2} implementation. For our ImageSubject, we have five key frames for each movie shot. So we also train the state-of-the-art  temporal models MEGA ~\cite{chen2020memory} and RDN \cite{deng2019relation} by the official implement of \cite{chen2020memory}. As the same inference process in their paper, temporal models ~\cite{chen2020memory, deng2019relation}  predict the subject for a single frame by using other remaining frames of the same shot as support images. All the methods are trained with the same backbone ResNet50 ~\cite{he2016deep} for fair comparison. All the backbone network weights are initialized by pre-trained models from ImageNet \cite{deng2009imagenet}.  We use the same input setting for all methods and use the hyper-parameter setting from their official optimal configuration. 

\begin{figure*}[]
\begin{center}
\includegraphics[width=1\linewidth]{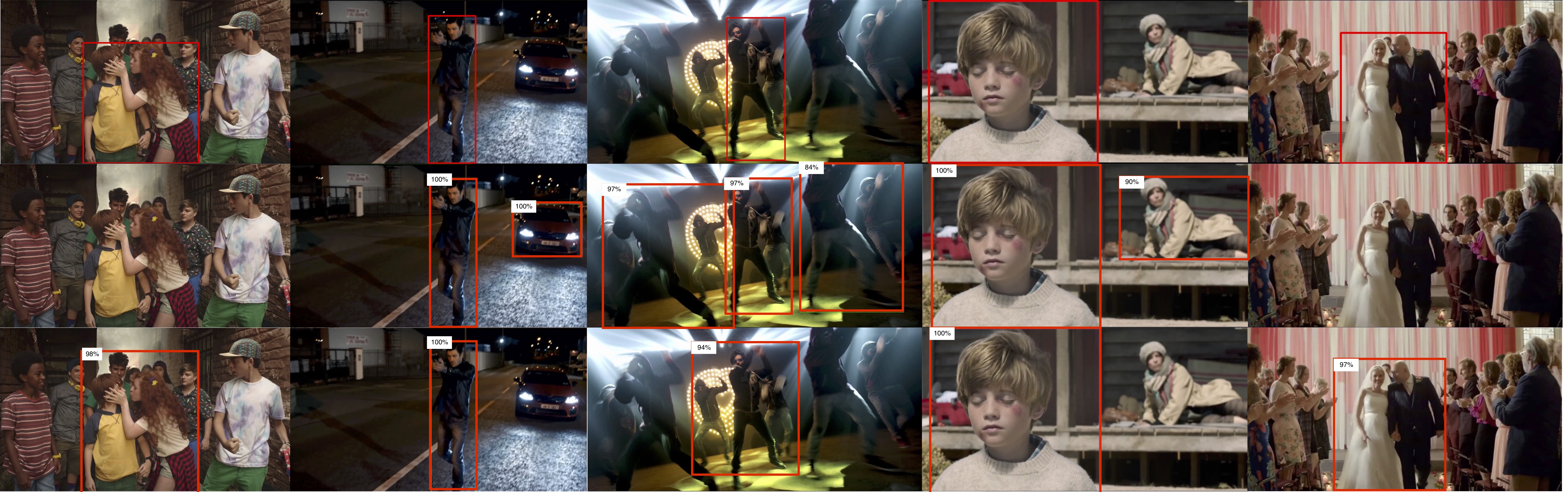}
\end{center}
\vspace{-0.3cm}
\caption{Examples for the subject prediction results by different methods. The first row is the ground truth for the testing image. The second row is the prediction results by the traditional initialization guess method Faster-RCNN. The third row is prediction results by the transformer based method \ac{DETR}. The prediction threshold is 0.9 for both methods.}
\label{fig:figure4}
\end{figure*}

\begin{figure*}[]
\begin{center}
\includegraphics[width=1\linewidth]{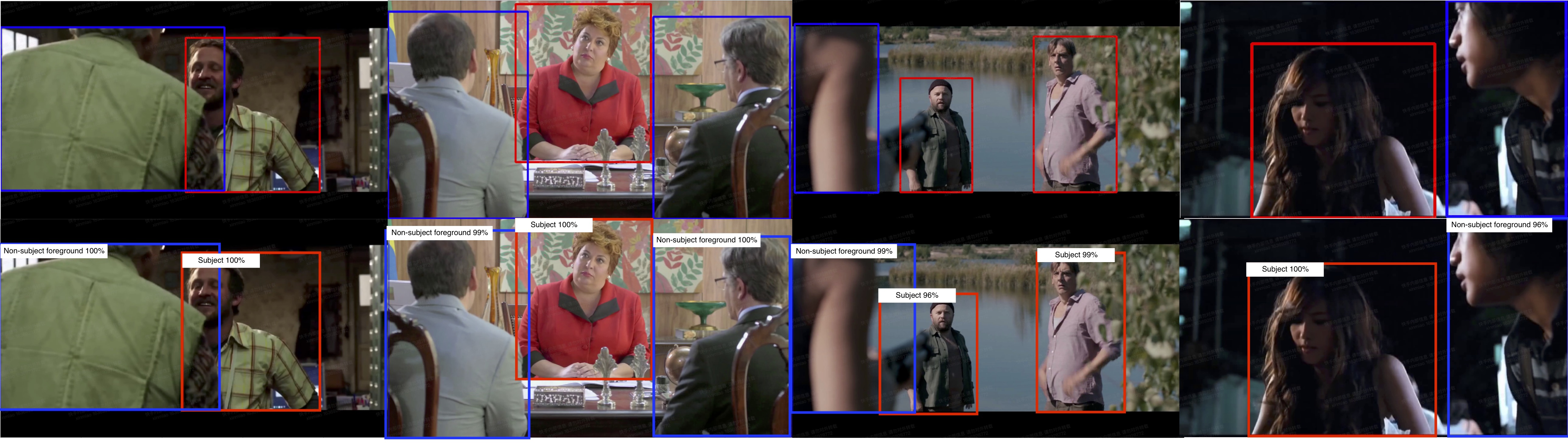}
\end{center}
\vspace{-0.3cm}
\caption{Detection results (second row) by \ac{DETR} for images have both subject and non-subject foreground object classes. The first row is the ground truth. The red bounding box is for the subject, the blue bounding box is for the non-subject foreground object. The prediction threshold is 0.9 for both methods.}
\label{fig:figure5}
\end{figure*}

\subsection{Subject Detection Results}
The subject detection results of all used methods are shown in \cref{tab:tabel3}. On the COCO dataset, Faster-RCNN is better than RetinaNet, but the result is the opposite on ImageSubject. Compared to RetinaNet, Faster-RCNN is a two-stage model, it predicts the result for each interest region independently. So it lacks global information, which is important for our subject detection. It explains why the two-stage model does not perform well in our task. RetinaNet predicts results on the whole feature map instead of the \ac{ROI}-aligned small feature map~\cite{he2017mask}, so it has a larger view field than the one in the Faster-RCNN architecture. That is why RetinaNet performs better than Faster-RCNN on ImageSubject. \ac{DETR}~\cite{carion2020end} achieves the best performance on our ImageSubject and outperforms other previous methods by large margins. We can see that on COCO object detection, the most recent proposed Spare-RCNN \cite{sun2021sparse} gives the best results. However, it performs much worse than \ac{DETR} on our dataset. The reason is that Sparse-RCNN lacks the cross-attention between the object queries and the dense features, it cannot reason the context for the whole image. We conclude that the cross attention between object queries and dense features is important for subject detection. The gap in AP results on our dataset between Faster-RCNN \cite{ren2015faster} and \ac{DETR} is much larger than it's on the COCO dataset, which indicates that that the \ac{DETR}'s architecture indeed fits better for our subject detection task. As shown in \cref{fig:figure4}, \ac{DETR} can better understand the context of the scene and contrast the clearness, position, scale information and relations between objects to localize the subject in the image. In \cref{fig:figure5}, we show the subject detection results by \ac{DETR} for images that have non-subject foreground object besides the subject, it indicates \ac{DETR}'s ability to understand spatial information in images. \ac{DETR} also offers the fastest inference speed on ImageSubject (see \cref{tab:tabel3}). 

In this paper, we offer the baseline result and believe there is still a lot of space to improve on ImageSubject. For example, deformable \ac{DETR} ~\cite{zhu2020deformable} can be used to better detect small subjects on ImageSubject.

MEGA~\cite{chen2020memory} and RDN~\cite{deng2019relation} are the state-of-the-art methods for video object detection on benchmark ImageNet VID~\cite{russakovsky2015imagenet}. However, they perform badly on our dataset. It is due to the ImageNet VID mostly containing few objects for each video, their hard cases are the object with motion blur, occlusion or out of focus in the single frame. These methods try to find the clues from other frames, which is different from our dataset. Specifically, we need to better understand the scene and mining the relations on ImageSubject. We believe that designing temporal model for subject detection is a research direction for future work too. For example, sometimes we need the motion information of objects to identify the subject.

\begin{figure*}[t]
\begin{center}
\includegraphics[width=1\linewidth]{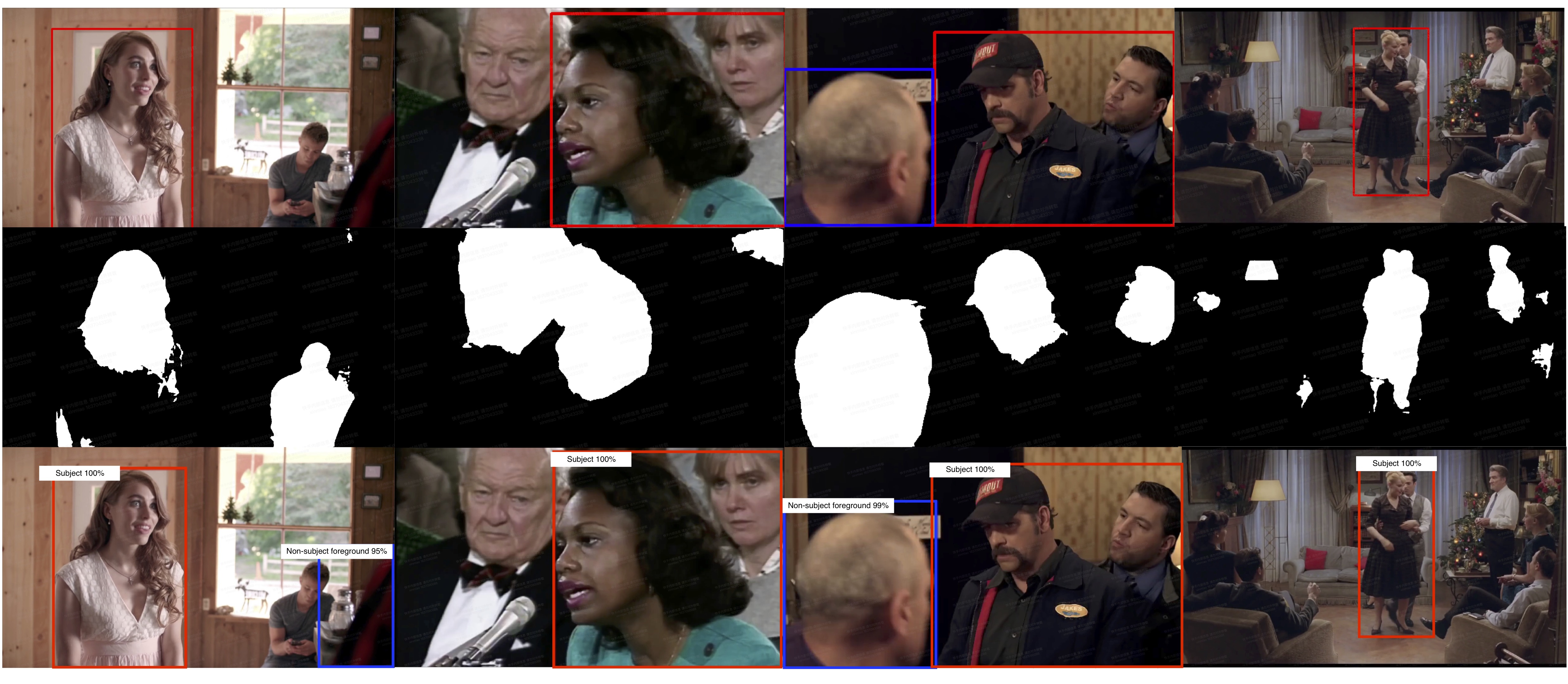}
\end{center}
\vspace{-0.3cm}
    \caption{Results of salient object detection model and subject detection model\cite{deng2018r3net} on our ImageSubject. The first row is the labeled data in our testing test. The red bounding box is for the subject, the blue bounding box is for the non-subject foreground object. The second row is the saliency map predicted by the salient object detection model. The third row is the subject detection result by using \ac{DETR}.}
\label{fig:figure7}
\end{figure*}

\begin{table}[h]
\begin{center}
\begin{tabular}{c|c|c}
\hline
Model& Train set & Testing AP on ImageSubject  \\
\hline
RetinaNet & ImageSubject  & 72.7 \\

\hline
RetinaNet& DUTS &  48.2 \\

\hline
\end{tabular}
\end{center}
\caption{Subject detection results for \ac{DETR} on our ImageSubject test by using different training sets. }
\label{tab:table4}
\end{table}

\begin{figure}[t]
\begin{center}
\includegraphics[width=1\linewidth]{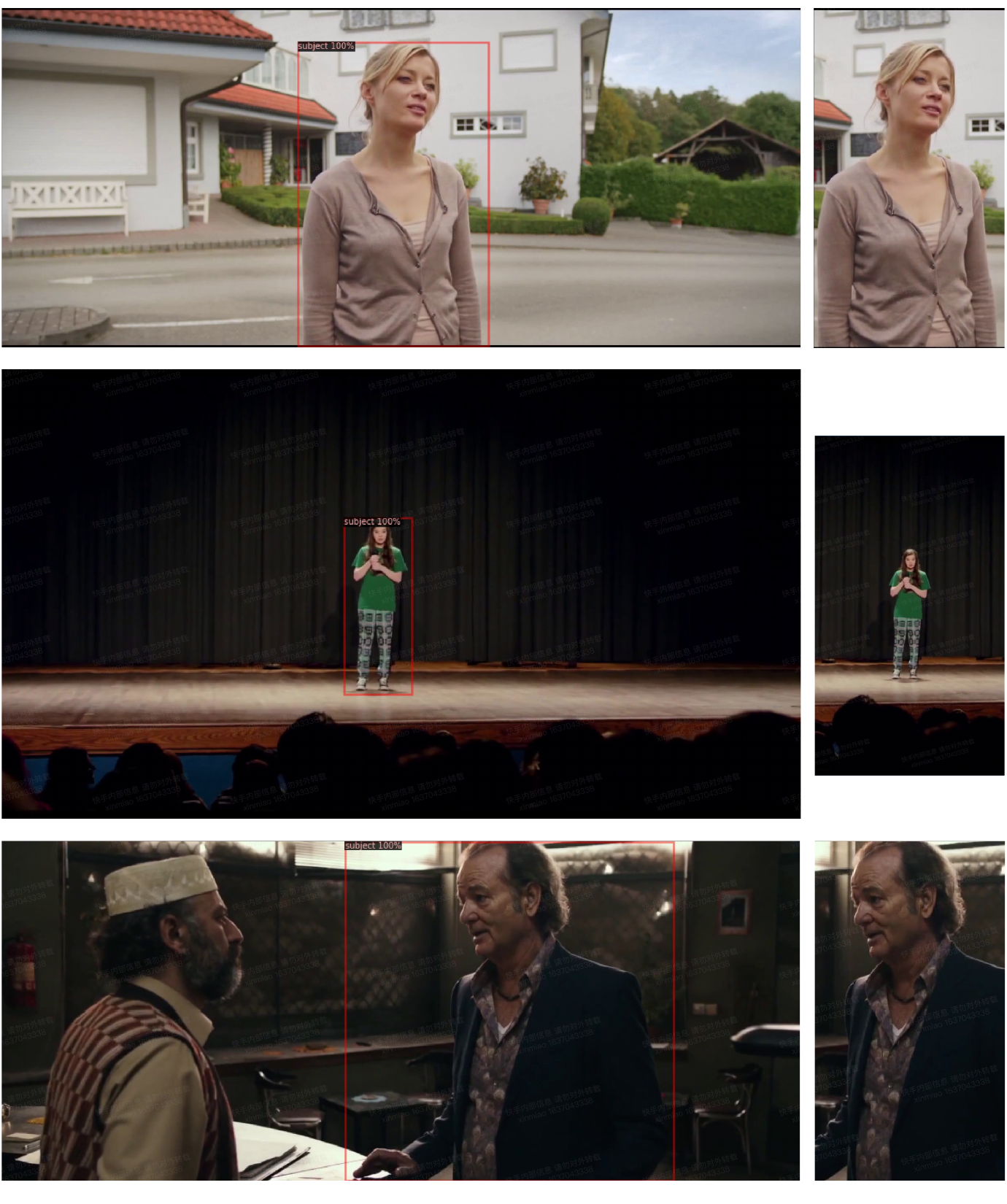}
\end{center}
\vspace{-0.3cm}
\caption{Image editing results. Left are the images from movie shots with different scales. Right are the cropped images that fit the phone screen by using the subject detection results. }
\label{fig:figure8}
\end{figure}

\subsection{Comparison with Saliency Detection Dataset}\label{sec:comparison_with_saliency_detection_dataset}
One of the related types of dataset is saliency detection. We conduct experiment to compare our ImageSubject with the large-scale saliency detection dataset DUTS \cite{wang2017learning}. Since the ground truth for salient object detection is segmentation mask. We generate the bounding box for the object from its mask. Then we train a RetinaNet detection model on datasets. For DUTS we use all the 10\,553 images in the training set. For ImageSubject, we random sample 10\,000 images from different shots as the training set. We compare their subject detection results on our ImageSubject testing set. We do not use the non-subject foreground object class of  ImageSubject in this part. In \cref{tab:table4}, we can see that the salient object bounding box detection model trained on DUTS performs badly on our testing dataset. Because the salient objects detection dataset usually contains few objects (mostly one) for each image. The model trained on it cannot handle the complex cases in our testing set.

We also test  state-of-the-art salient object segmentation model \cite{deng2018r3net} on our ImageSubject. This model~\cite{deng2018r3net} is pretrained on the  \ac{SOD} dataset MSRA10K \cite{cheng2014global}. In \cref{fig:figure7}, we can see that the subject that the photographer wants to highlight may not be the most salient object in the image detected by the salient object segmentation model~\cite{deng2018r3net}. The model needs to understand the context and the relations between objects to localize the subject.  The model trained on the salient object detection dataset can not handle the subject detection task. That is why our proposed ImageSubject is unique.

% Due to the image resolution of our dataset if larger than DUTS, we downsize our images to be comparable with DUTS.

\section{Application}
We believe there is a wide range of potential applications for subject detection. It can be a useful prepossessing to help other tasks like video compression, image editing, image quality assessment, image retrieval, etc. In this part, we show one application of image editing with the help of our subject detection model.

As the short-video become more popular nowadays, the user usually watches the full screen image or video on their cell phones. However, the images or videos are created with different aspect ratios. In order to make images fit the phone screen, we need the subject detection model to localize the main subject in the image then we can do the image editing automatically~\cite{gleicher2008re}. The image editing results can be found in \cref{fig:figure8}. Images on the left are from movies with different scales. If we put them on the phone directly, the subjects may lack details for the viewers. We crop the main subject to fit the cell phone screen scale by using our subject detection results. The cropped results are shown on the right of \cref{fig:figure8}. These application results demonstrate the importance of subject detection.

\section{Conclusion}
In this paper, we introduce the first large-scale dataset ImageSubject for object detection . We collect the dataset with strong diversity, it contains 107\,700 images from 21\,540 movie shots. Two classes' bounding boxes are  professionally annotated: subject and non-subject foreground object. ImageSubject has the spatial, scene, and temporal information for use. Our dataset helps the model localize the subject in the image that the photographer wants to highlight.  As demand for understanding images and videos increases in this era, we believe our proposed  dataset  can  help  researchers  develop  better techniques in this direction.

Different detection models were implemented on our ImageSubject. We analyze the relationship between model architectures and their performance on ImageSubject. We give the baseline result by using the popular transformer-based model \ac{DETR}. We believe there is still a lot of space to improve for future work, not only for single frame detection models but also for temporal subject detection models.

We conduct detailed analysis for the proposed dataset and also implement experiments to show the difference between our subject detection and other related tasks like saliency detection. We prove that the existing dataset can not solve the proposed  subject detection task well. It indicates the importance of the proposed dataset ImageSubject. We believe our proposed dataset can be useful for many other tasks.

{\small
\bibliographystyle{ieee_fullname}
\bibliography{egbib}
}
\end{document}